\documentclass[A4, 10 pt, conference]{ieeeconf}  
\IEEEoverridecommandlockouts   
\usepackage{wrapfig}
\usepackage{cite}
\usepackage{algorithm} 
\usepackage{algorithmic}  
\usepackage[algo2e]{algorithm2e} 
\usepackage{amsmath,amssymb,amsfonts}
\usepackage{algorithm}
\usepackage{graphicx}
\usepackage{textcomp}
\def\BibTeX{{\rm B\kern-.05em{\sc i\kern-.025em b}\kern-.08em
    T\kern-.1667em\lower.7ex\hbox{E}\kern-.125emX}}
    
    \usepackage{lineno}
\usepackage{lettrine}

\usepackage{url}
\usepackage{moreverb}

\usepackage{graphicx}
\usepackage{subcaption}
\usepackage{float}
\usepackage{amsmath}
\usepackage{array}
\usepackage{multirow}
\graphicspath{ {Figures/} }

\usepackage{scalerel}
\usepackage{tikz}
\usetikzlibrary{svg.path}

\definecolor{orcidlogocol}{HTML}{A6CE39}
\tikzset{
  orcidlogo/.pic={
    \fill[orcidlogocol] svg{M256,128c0,70.7-57.3,128-128,128C57.3,256,0,198.7,0,128C0,57.3,57.3,0,128,0C198.7,0,256,57.3,256,128z};
    \fill[white] svg{M86.3,186.2H70.9V79.1h15.4v48.4V186.2z}
                 svg{M108.9,79.1h41.6c39.6,0,57,28.3,57,53.6c0,27.5-21.5,53.6-56.8,53.6h-41.8V79.1z M124.3,172.4h24.5c34.9,0,42.9-26.5,42.9-39.7c0-21.5-13.7-39.7-43.7-39.7h-23.7V172.4z}
                 svg{M88.7,56.8c0,5.5-4.5,10.1-10.1,10.1c-5.6,0-10.1-4.6-10.1-10.1c0-5.6,4.5-10.1,10.1-10.1C84.2,46.7,88.7,51.3,88.7,56.8z};
  }
}

\newcommand\orcidicon[1]{\href{https://orcid.org/#1}{\mbox{\scalerel*{
\begin{tikzpicture}[yscale=-1,transform shape]
\pic{orcidlogo};
\end{tikzpicture}
}{|}}}}

\usepackage[draft]{hyperref}
\usepackage{float}
\floatstyle{plaintop}
\restylefloat{table}

\begin{document}
\title{\LARGE \bf
INTENT: An LSTM Framework for Vehicle Intention Prediction in Intersection Scenarios with Comprehensive Ablation Analysis}
%Development of the INTVIP: Intersection Vehicle Intention Prediction Pipeline}

\author{Logine M. Zaki$^{1,2}$ and Catherine~M.~Elias$^{1,2\orcidicon{0000-0002-1444-9816}\,}$,% 
\thanks{*This work was not supported by any organization}% <-this % stops a space
\thanks{$^{1}$C-DRiVeS Lab: Cognitive Driving Research in Vehicular Systems, Cairo, Egypt
{\tt\small cdrives.researchlab@gmail.com}}%
\thanks{$^{2}$Computer Science and Engineering Department - Faculty of Media Engineering and Technology - German University in Cairo, Egypt}%
\thanks{{\tt\small logine.elkelani@student.guc.edu.eg, catherine.elias@ieee.org}}%
}

\markboth{Journal of \LaTeX\ Class Files,~Vol.~14, No.~8, August~2015}%
{author1 \MakeLowercase{\textit{et al.}}:title here}



\maketitle
\begin{abstract}
Vehicle intention prediction is a pivotal aspect in the agility  and safety of autonomous vehicles in all driving scenarios; if  genuine  enhancement of autonomous vehicles are required, we need to make them adopt human interpretation of driver's intention especially in cases that require a lot of human interaction as well as complex driving behaviors  like the ones at intersections, roundabouts and emergency cases such as   sudden stops where vehicle intention prediction helps in taking the correct evasive action within a real time period where every second of action makes an impact  and can prevent a catastrophe from taking place. In the worst case, it helps minimize the damage  and make safety a priority. Intention prediction can also be used to enhance trajectory prediction (intention conditioned trajectory prediction). 
In this study, The INTENT framework is proposed using LSTM model to predict the vehicle's intention at intersections 2 seconds ahead of the event occurrence to predict whether the cars in intersections are going straight, turning left, or turning right. Various model experiments and ablation study are thoroughly tested on InD dataset achieving 99.71\% accuracy. 
%that utilised  the opendrive maps provided in order to first allocate which road the vehicle is in and then another algorithm for road transitions that is used to get the ground truth before feeding input features to the model.
\end{abstract}


% \fi
%\IEEEpeerreviewmaketitle
\section{Introduction and Related Work}\label{sec1}
In continuous efforts of enhancing road safety to reduce accidents , autonomous 
 vehicles could be a leading participant if it is able to avoid main causes of accidents like human error, having a misleading intention , taking the wrong action ,  unexpected lane changes, vague  driving behaviors at intersections and sometimes not seeing vehicles if they are occluded.Understanding the scene and even predicting the intention of the target vehicle could alert the driver if the driven car is not fully autonomous or even better, if it is fully autonomous, would take the safest evasive action to ensure optimal safety. Intention prediction is vital for all road users for example: pedestrians,cyclists and vehicles. \par
For each road user, there are even various traffic and road scenarios like pedestrian intention prediction at intersections and highways .In this paper 
 ,vehicle intention prediction is going to be our main focus. Car accidents at intersections  suffer fatality and severity  for few reasons like:vehicles could be at high speed, the collision angle could be  deadly ( 90 degrees ) ; imagine collision of 2 cars one  that is crossing from South to North and another one from East to West) which reinforces the vitality of  addressing the challenge of vehicle intention prediction at intersections which could exceedingly improve   road safety and accelerate  the development process of autonomous vehicle. This  is why vehicle intention prediction at intersections is    our paper motivation.
 %%%%%%%%%%%%%%%%%%%%%%%%%%%%%%%%%%
 \begin{comment}
     \begin{figure}

 \centering
 \vspace{-20pt}
 \includegraphics[width=0.4\textwidth]{Figures/intersection.png}
 \caption{An illustration of possible driving maneuvers of the target vehicle where the target vehicle is the red one and the ego vehicle is the green one.}
 %of an EV(ego vehicle) that needs to predict the intention of the leading TV (target vehicle) in order to take the correct action, maximize safety and avoid ambiguity of determining the driving behaviour of the target vehicle which might lead to a catastrophic collision affecting even any other vehicles behind the ego vehicle if the EV hinders the traffic flow of the road in the East.}
 \label{intersection}
\end{figure}
 \end{comment}

Vehicle intention prediction is a dominant  factor in the agility  and safety of autonomous vehicles .Intention prediction is a classification problem thus diverse classifiers throughout the years  have been exploited to carry out intention prediction where some are categorised as machine learning techniques such as support  vector machines (SVM) \cite{svm}  and hidden Markov models (HMM) \cite{HMM}.

The other category is deep learning models , that have been proven exceedingly significant   such as Recurrent Neural Networks (RNN) which are best utilised to model short sequences of data but are deficient in modelling long sequences due to the vanishing gradient problem , and that's why Long Short Term Memory (LSTM) is predominantly  used in using  the past sequential trajectory info of a  vehicle to be able to predict the intention overcoming the hindrance of  long sequences \cite{LSTM} which is highly used in pedestrian intention prediction \cite{manzour2024development} and so many researchers use the LSTMs for vehicles intention predictions \cite{abouras2025observation, amer2024enhancing}, yet it is observed that they mainly tackle this problem only on Highways.

Additionally, a lot of recent studies have been speculating  different variants of LSTM in vehicle intention prediction as in \cite{attentionLstm} where bidirectional LSTM coupled with attention mechanisms are used. Furthermore, some researchers started to use other algorithms like the Knowledge graphs embeddings with Bayesian Inference in the highway vehicles intentions predictions \cite{manzour2026design, 11376512} and transformers architecture \cite{abdelshafy6234830caution, elkammar2026vit} but also on pedestrian intention prediction.

From this conducted literature, it has been observed that vehicles intention predictions in intersections are under investigated using the different architecture especially due to the lack of datasets. Also, there is no proper ablation study to reach high accuracies in intersections scenarios.
Accordingly, this work presents the INTENT Framework. This framework tackles the vehicle intention prediction model using LSTM architecture, specifically in intersection scenarios that contribute to the safety of autonomous vehicles. In addition, the paper provides an ablation study that aims at architecture with high accuracy.
\begin{figure*}[t]
    \centering
    \includegraphics[width = 0.7\linewidth]{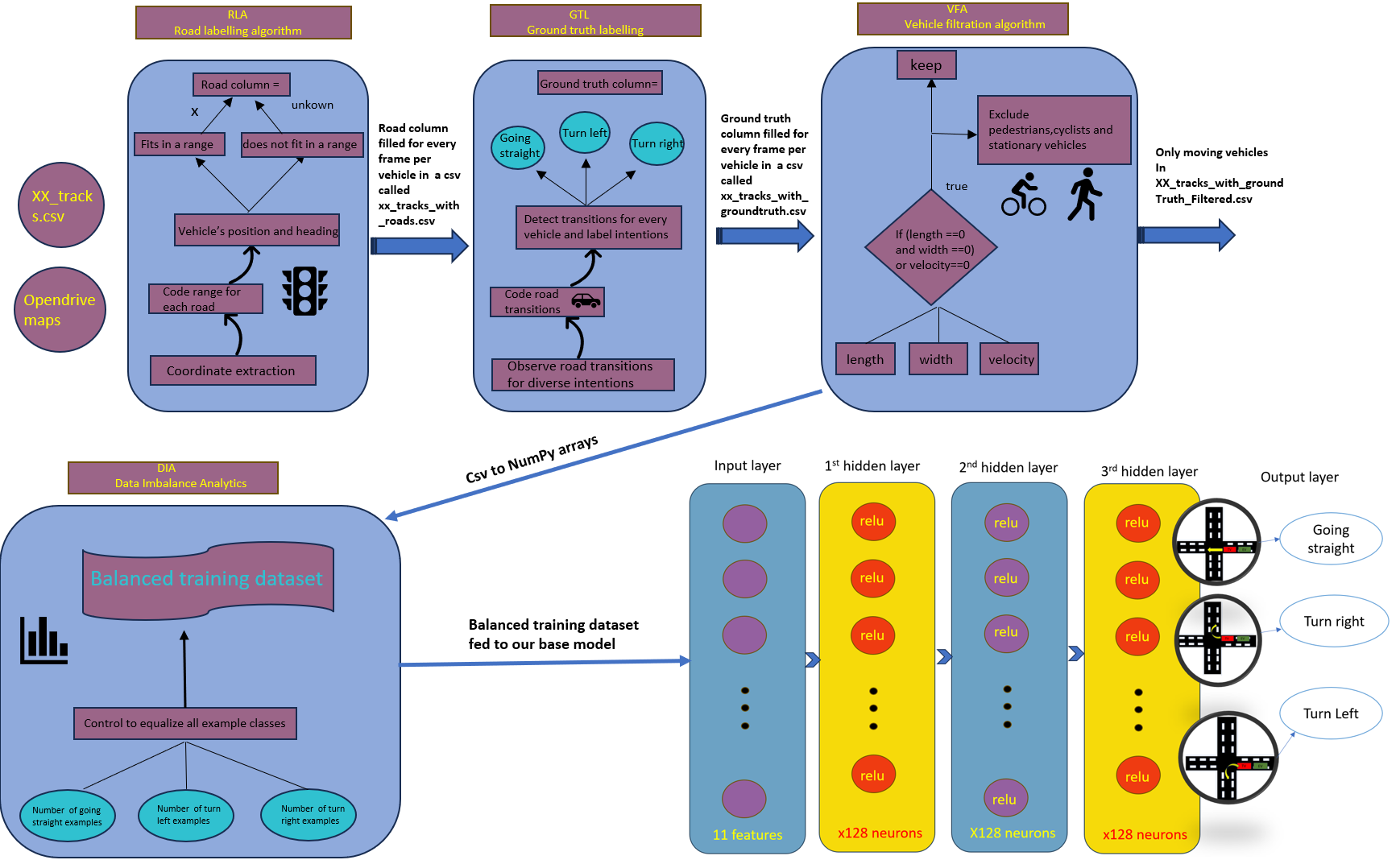}
    \caption{The INTENT Pipeline Stages Overview}
    \label{INTENT}
\end{figure*}
\section{INTENT Pipeline Implementation}
\vspace{-7pt}\subsection{INTENT Pipeline Overview}
Fig. \ref{INTENT} illustrates the implemented detailed pipeline overview. The INTENT Pipeline consists of 2 main stages; The Data Preparation Stage including the Road Labeling, Ground Truth Labeling, and Vehicle Filtration, followed by the prediction model stage which includes the Data Imbalance Analysis and the Model Implementation.
\vspace{-7pt}\subsection{The Data Preparation}\label{sec3}

As far as we are concerned and from the thorough search we carried out on a long period of time  a huge research gap that we faced is that there was not any ego vehicle view dataset capturing driving behaviours at intersections so we decided to work with BEV datasets and leverage the advantages of not being constrained by sensor occlusions or limited ranges .Not only was it decided  to use BEV , but also it was planned that the model that will be built would rely on features that could be extracted regardless the view thus it would be general enough and view independent  thus easily applicable on multiple views and scenarios.

\begin{figure}[H]
    \centering
    \includegraphics[width = 0.7\linewidth]{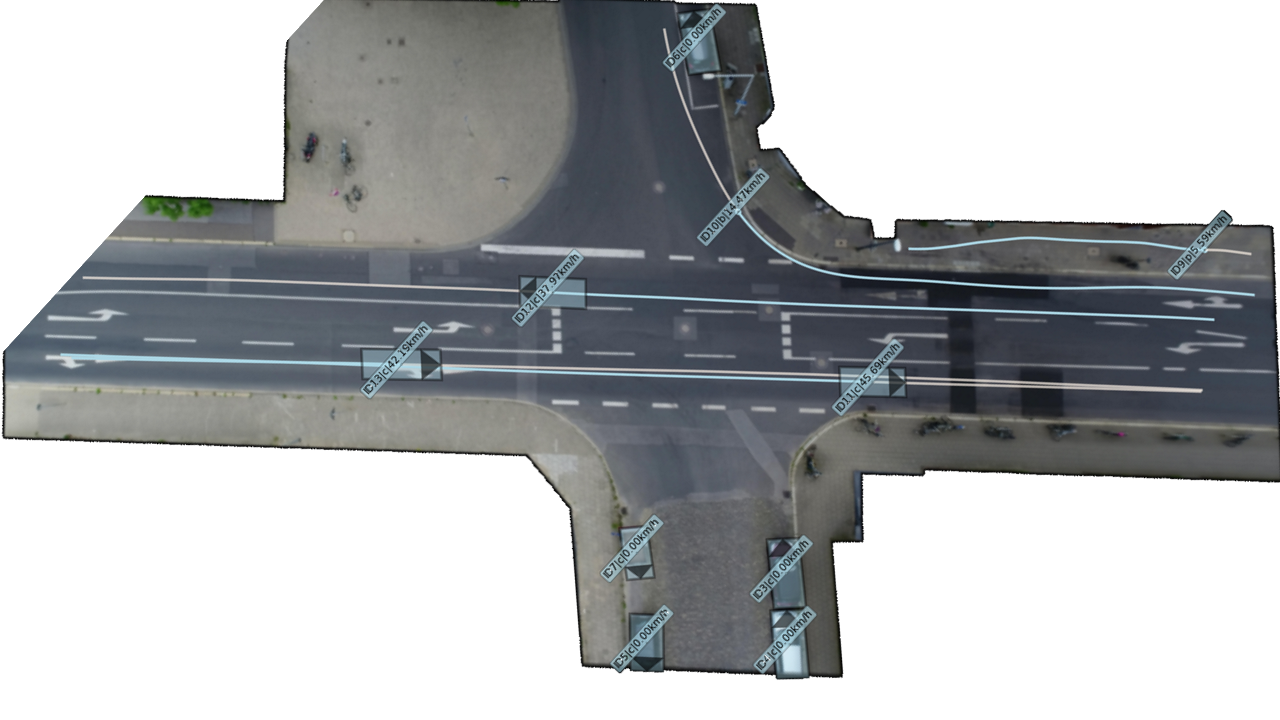}
    \caption{Visualizer tool offered by the inD dataset to observe vehicle behaviour at different frames }
    \vspace{-10 pt}
    \label{VisualizerTool}
\end{figure}

InD \cite{inDdataset} which is a BEV dataset capturing different driving behaviours at different intersection structures (T intersection, cross-intersection,etc..) as in Fig. \ref{VisualizerTool} was used.
\vspace{-10pt}\vspace{-7pt}
Unfortunately,  a problem was faced that  no ground truths were available alongside the dataset trajectories  so labelling mechanism is developed as follows.

\begin{figure}[H]
    \centering
    \includegraphics[width = 0.7\linewidth]{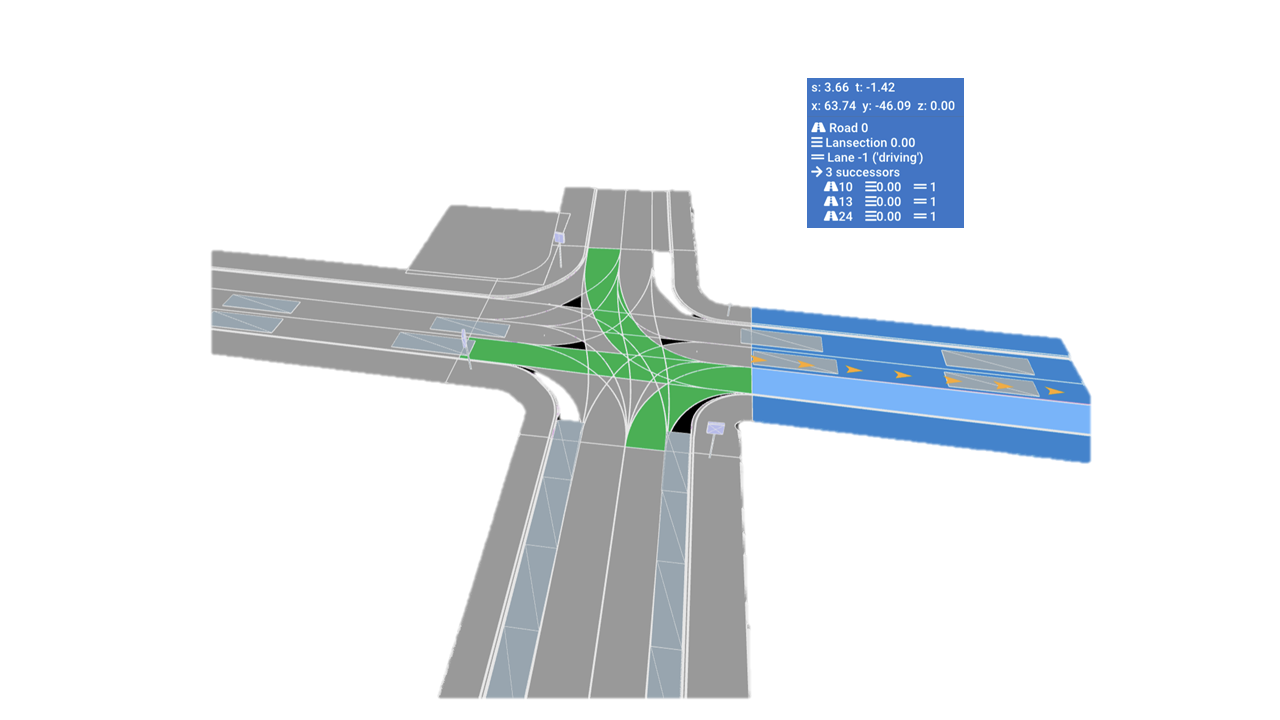}
    \caption{ a cross intersection open-drive map   to extract the  x and y coordinates of different roads}
    \label{opendrive map}
\end{figure}

First the Road Labelling Algorithm (RLA), Using open-drive maps that  were  included in the dataset  as shown in Fig. \ref{opendrive map}  when opened (by online open drive viewer ) , through  moving the  cursor  over the maps ,the x and y coordinates are yielded thus manual coordinate extraction was done for every road .  As a preliminary step,  all lower and upper limits of x and y coordinates of all roads   were coded  in order to label the road the  vehicle is in at each frame  based on the x and y coordinates of the vehicle's position . Solely using  x and y coordinates ,  common areas between roads were found so  another property was needed in order to distinguish between roads leading  to the inclusion of the heading  angle  which successfully  identified  between roads. 
    
%%%%%%%%%%%%%%%%%%%%%%%%%%%%%%%%%%
\vspace{-10pt}\vspace{-10pt}
%%%%%%%%%%%%%%%%%%%%%%%%%%%%%%%%%%
Common areas between roads still existed even after heading inclusion  so a common range was extracted and the road column  value was coded as “either road x or road y “ till the vehicle moved to a unique range linked  to a single road.

A question could come to mind and the question might be how is  the heading angle  extracted when all the open-drive map offers are the x and y coordinates? Using the visualizer tool ,for example, if we wanted to extract the heading of a certain road , we observed different vehicles that drove in the same road and getting their ids , we refer to the csv file and observe their heading angles and extract a common range that fits all vehicles driving on each road .

The inclusion of the heading angle  was done for all roads not just the ones that had overlaps. There was an unknown value filled in the column of road if something did not fit the range or a vehicle was not obeying the lane markings or allowed directions.

The second step after the road column was filled for each frame is the Ground Truth Labeling (GTL) where road transitions were coded to label events / ground truth behaviour as either going straight , turn left or turn right . Afterwards, it was manually validated and revised against the visualizer tool to make sure the ground truth matches the real action the vehicle took.

The third and last step is the implementation of the Vehicle Filtration Algorithm (VFA) which is another filtration algorithm written to filter out and exclude tracked objects of classes pedestrians, parked vehicles/ stationary vehicles and cyclists in order to merely focus on  vehicle intention prediction for  moving vehicles.
\vspace{-7pt}\subsection{Features}\label{sec4}
% One point that was taken into consideration is to pick datasets that proffer general features  not related to traffic rules  in order not to constraint us in  certain environments  so numerical features from vehicle trajectories were provided. 
%and no features related to traffic rules or road structures were included  as these widely vary  from one location to another  and 
Below are the main features used: the vehicle’s center positions and heading angle  $[P_x, P_y,   \theta]$, vehicle velocity and acceleration $[v_x,  v_y,  a_x,  a_y]$, longitudinal and lateral velocity and acceleration $[v_{lon},  v_{lat},  a_{lon},  a_{lat}]$, and the vehicle dimensions including the length $l$ and width $w$.

%These features help support  broad applicability  where it can  simply be tested on different environments provide that  these features are existent ; for example it  can be tested  on simulator environments such as Carla by placing sensors that extract those features for different vehicles and feeding it to our model. 
\vspace{-7pt}\subsection{Model}\label{sec5
}
\begin{comment}
 \subsubsection{Specs and tools}
 Our model proposal  is an LSTM deep learning model in order to exploit sequential nature of  the frames and numerical features of each vehicle . Tensorflow and keras api were used and our model was coded in visual studio code . The hardware used to run the model is a laptop with an  i9-14900HX cpu  and nvidia’s geforce rtx 4080.The intention prediction of a vehicle is treated as a classification problem where the output consists of 3 classes which are going straight , turn left and turn right.
\end{comment}
 \subsubsection{Input preparation}
 For the inputs a subset of the  features mentioned in the aforementioned features section were used , but initially they were in the form of a csv , so an input preparation algorithm  was written to transform filtered csv (csv excluding pedestrians ,cyclists and stationary vehicles) into 3d numpy arrays to be fed to the model along with  a one dimensional labels array  (from the ground truth column mentioned earlier ) to be input as the ground truth of the model.

\subsubsection{Controlling imbalance}
Using pandas, an analytical part was run to investigate the number of input examples from  each class from going straight , turn right and turn left. The justification  behind the analytics part is  to avoid data imbalance. Our doubts were validated and the going straight input examples were exceeding   the other 2 classes so a counter was used and only a number equal to the examples of the other 2 classes
(turn right and turn left ) were input as examples of going straight.\par

\subsubsection{Details of input reshaping}
InD dataset has a  frame rate of 25 frames  per second so in our model  a 20 frame window size was used  and  padding was done  for vehicles that had less than 20 frames before the event happened . Our model predicts the intention of the target vehicle 2s   before the action occurs so as per the frame rate mentioned above , in data preparation   an algorithm was written such that for every vehicle in the filtered csv , it gets the row number where the action happens and goes back 50 rows (1 row = 1 frame ) and gets the row  number of the first frame for each vehicle and then does feed sequences of 20 frames (aligned with the ground truth in the one dimensional labels array)  as a single example till it reaches the end row (which is the row the action happened - 50 rows).
\subsubsection{Model architecture}
An input layer with a combination of the features mentioned in the features section was used  and further details are to  be further discussed in  the experiments section.Then our model consists of 3 hidden layers each layer has 128 neurons and all 3 hidden layers use the RELU activation function as our base model. Finally , the output layer has 3 output nodes and uses soft-max function in order to limit output between 0 and 1 acting as   a probability like one where the class with the highest  probability is considered the output belonging to this class/intention in our case.

Batch Normalisation is used between the hidden layers.As kpis/evaluation metrics accuracy , precision  recall,f1-score and loss  are used.Our dataset is split as 80 percent training data, 20 percent validation data and for test data set  a separate recording at the same location was used to further evaluate the model.

The categorical cross entropy was used as a loss function and as an optimization function Adam optimizer was used with a learning rate of 0.001.A batch size of 32 and 50 epochs were used.

\begin{table*}[t]
    \centering
    \caption{Results of Experiment 1 comparing the 4 different models}
    \label{Experiment 1}
    \begin{tabular}{c||c|c|c|c}
        \hline
        Metric & 5-Feature Model & 7-Feature Model & 11-Feature Model & 13-Feature Model \\
        \hline
        \hline
         Test Accuracy  & 0.8264 & 0.87 & \textbf{0.9151} & 0.8288\\
        \hline
         Test Precision & 0.8268 & 0.8705 & \textbf{0.9151}& 0.8331 \\
        \hline
        Test Recall & 0.8264 & 0.8700 & \textbf{0.9146} & 0.8278 \\
        \hline
        Test F1 Score &  0.8254 & 0.8698 & \textbf{0.9150} & 0.8294 \\
        \hline
        Test Loss &   0.5811 & 0.7418 & \textbf{0.7214} & 1.4496 \\
        \hline
    \end{tabular}
\end{table*}
\begin{table*}[t]
    \centering
    \caption{Results of Experiment 2 comparing all model variants with respect to the base model}
    \label{Experiment 2}
    \begin{tabular}{p{1.5cm}||p{1.5cm}|p{1.5cm}|p{1.5cm}|p{1.5cm}|p{1.5cm}|p{1.5cm}|p{1.5cm}|p{1.5cm}}
        \hline
        Metric & base model & Regularization& bidirectional  & 25 window   & 2 hidden layers& 4 hidden layers& 0.01 learning rate & 1.5s prediction window  \\
        \hline
        \hline
         Test Accuracy  & \textbf{0.9151} &  0.9146 & 0.8443 & 0.8923 
         & 0.8909
         & 0.8555
         &0.5587
         &0.8749\\
        \hline
         Test Precision & \textbf{0.9151} & 0.9160 & 0.8457& 0.8941 
         &  0.8917
         &0.8570
         & 0.5650
         &0.8766\\
        \hline
        Test Recall & \textbf{0.9146} & 0.9146 &  0.8429 & 0.8923
        &0.8904
        &0.8545
        &0.5291
        &0.8715\\
        \hline
        Test F1 Score &  \textbf{0.9150} &  0.9142 & 0.8438 & 0.8930
        &0.8906
        &0.8554
        &0.5246
        &0.8746\\
        \hline
        Test Loss &   \textbf{0.7214} & 0.5267 & 0.9669 & 0.6291
        &1.8022
        &0.9263
        &40604.2773
        &1.1998\\
        \hline
    \end{tabular}
\end{table*}
\section{Experiments and Results}\label{sec6}
In this section , model experiments that have been speculated is going to be clearly discussed ,categorised and their results are going to be compared respectively on training,validation and test datasets.Additionally, the performance graphs are going to be attached for some of the experiments for visualising the  results.
\vspace{-7pt}\subsection{Experiment 1: Testing different features}
The main aim of this experiment is to test the efficacy of using different number of input features on  our base model and  on the model accuracy, precision, recall, F1-score and loss. In this experiment, 4 different variations have been created using 5, 7, 11, and 13 features respectively. The input features of the models respectively  are $[P_x, P_y,   \theta,v_x, v_y]$,$[P_x, P_y,   \theta,v_x, v_y,a_x, a_y]$,
$[P_x, P_y,   \theta,v_x, v_y,a_x, a_y,v_{lon}, v_{lat}, a_{lon}, a_{lat}]$ and$[P_x, P_y,   \theta,v_x, v_y,a_x, a_y,v_{lon}, v_{lat}, a_{lon}, a_{lat},$l$,$w$]$.
%%% mention here which features are used with symbols in each model

The Architecture used for the 4 variations is unified for the sake of the comparison and any different parameter would be solely mentioned in each experiment accordingly. The unified architecture is an LSTM model with 3 hidden layers using relu activation functions and an output layer which uses softmax activation function  for  3 output nodes which are going straight,turn left and turn right .The optimizer used is Adam optimizer with the categorical cross entropy  loss function and a learning rate of 0.001. Batch normalization was used.The batch size is set to 32 and epochs to 50 where our sequence length is 20 frames ,our step size is 12 frames and our prediction window is 2s before the event occurs.

Table \ref{Experiment 1} illustrates that the  the 11 feature model yielded the best outcome thus we proceed with our upcoming experiments applying all  variations on  the model but in reference  with   the 11 feature model as our base model  and comparing every variant with the base model. 
%%% what is the best model --> 11 feature that's why we will continue with it.
\vspace{-7pt}\subsection{Experiment 2: Other model variants}
In the upcoming experiments, all model details of the base model are invariant and any modification will be solely mentioned in each experiment respectively. 
\subsubsection{Adding regularization layers}

\begin{figure}[H]
    \centering
    \includegraphics[width = 0.7\linewidth]{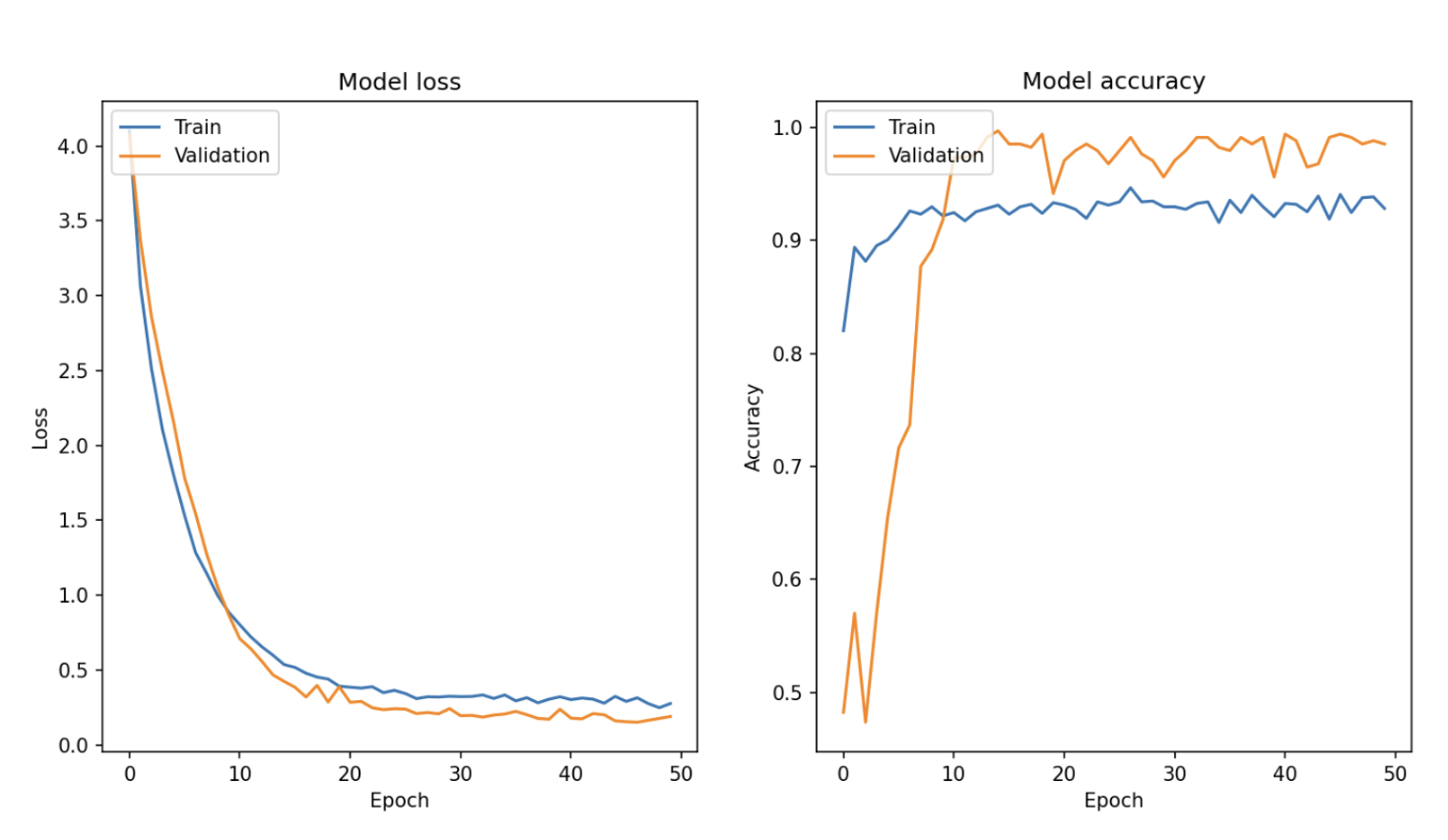}
    \caption{accuracy and loss  of training and validation datasets across epochs of the regularized model}
    \label{regular}
\end{figure}

Regularization and dropout layers of parameter 0.1  are added to help reduce over-fitting if there exists any yielding a maximum training accuracy of 0.9540, Precision of  0.9559, Recall of 0.9527, and loss of 0.2758. The maximum validation accuracy is 0.9971, val Precision is 1.0000, val Recall is 0.9912, and val loss is 0.4245. Fig. \ref{regular} shows graph results and Table \ref{regular_tab} shows the numerical results.
\vspace{-10pt}\vspace{-10pt}
\vspace{-10pt}\begin{table}[H]
    \centering
    \caption{Results of regularized model  on test dataset}
    \label{regular_tab}
    \begin{tabular}{|c|c|c|c|c|}
        \hline
         Class& Precision & Recall &  F1-score & support \\
        \hline
        Going Straight &  0.88   &   0.95   &   0.91   &    719\\
        \hline
        Turn Left &  0.94   &   0.93   &   0.94     &  857 \\
        \hline
        Turn Right &   0.94   &   0.83   &   0.88   &    486 \\
        \hline
    \end{tabular}
\end{table}\vspace{-7pt}
\subsubsection{ Bidirectional LSTM}
\begin{figure}[H]
    \centering
    \includegraphics[width = 0.7\linewidth]{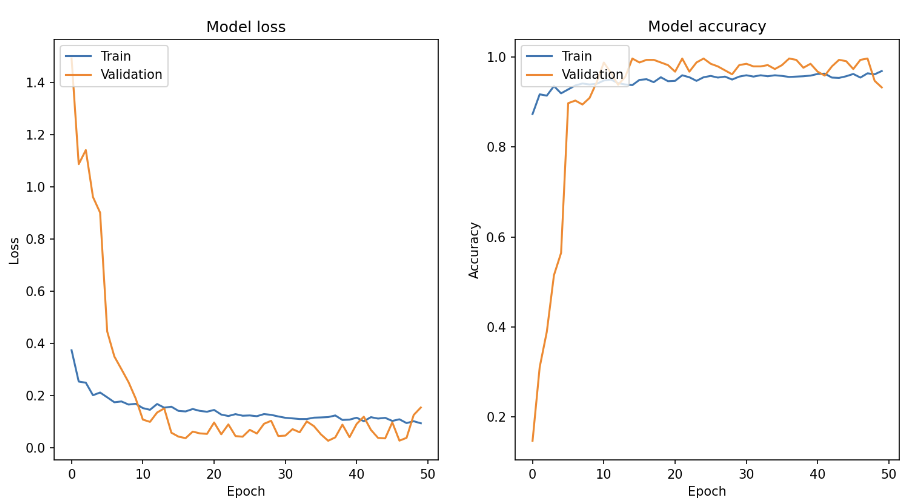}
    \caption{accuracy and loss  of training and validation datasets across epochs for bidirectional LSTM }
    \label{bidirectional}
\end{figure}

 The first hidden layer of the base model  is made bidirectional  to testify the effect of a bidirectional model and its impact on the results.  A maximum training  accuracy of 0.9727 ,Precision of  0.9727 Recall of  0.9725 and loss of 0.0795 .The maximum validation accuracy is 0.9971 , val Precision is 0.9971 ,val Recall is  0.99712  and val loss is 0.0266 were yielded. Fig.  \ref{bidirectional} shows graph results and Table \ref{bidirectional_tab} shows the numerical results.
\vspace{-10pt}\vspace{-10pt}
\vspace{-10pt}\begin{table}[H]
    \centering
    \caption{Results of bidirectional LSTM  on test dataset}
    \label{bidirectional_tab}
    \begin{tabular}{|c|c|c|c|c|}
        \hline
         Class& Precision & Recall &  F1-score & support \\
        \hline
        Going Straight &   0.78 &     0.96 &      0.86    &   719\\
        \hline
        Turn Left &  0.94   &   0.79  &    0.85 &      857 \\
        \hline
        Turn Right &   0.82     & 0.78  &    0.80   &    486\\
        \hline
    \end{tabular}
\end{table}\vspace{-7pt}
\subsubsection{Trying different activation functions}
\begin{figure}[H]
    \centering
    \includegraphics[width = 0.7\linewidth]{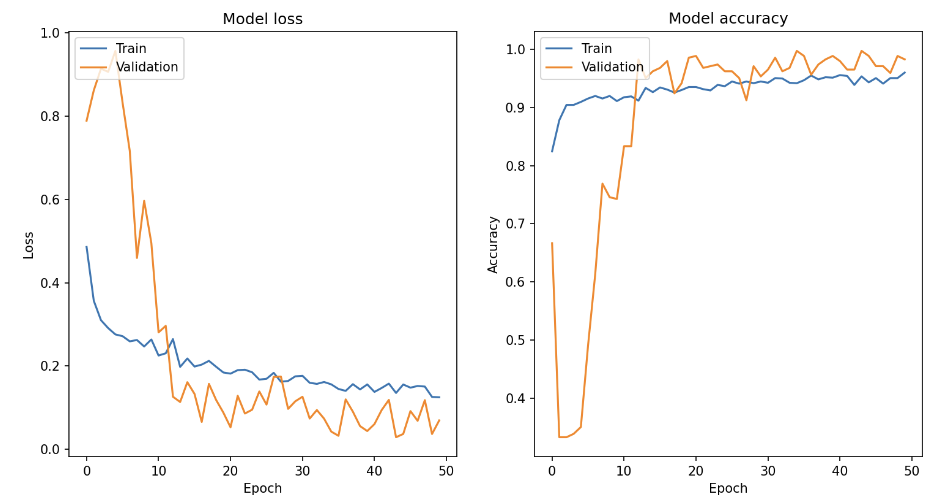}
    \caption{accuracy and loss  of training and validation datasets across epochs of using sigmoid as an activation function}
    \label{sigmoid}
\end{figure}

\begin{figure}[H]
    \centering
    \includegraphics[width = 0.7\linewidth]{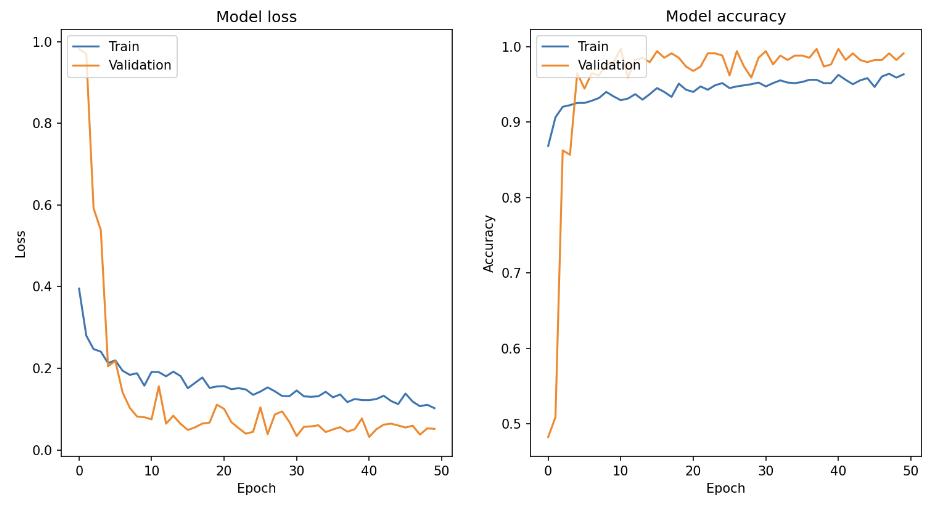}
    \caption{accuracy and loss  of training and validation datasets across epochs of using tanh as an activation function}
    \label{tanh}
\end{figure}

The 3 hidden layers of the LSTM is set once as sigmoid and  once as tanh  to testify the influence  of hidden layers' activation function choice on results. Our base model which used relu has the highest output  , the  the tanh and lastly the sigmoid has the lowest output but they are all slight differences (1 percent at maximum). Figures \ref{sigmoid} and \ref{tanh} show graph results and Tables \ref{sigmoid_tab} and \ref{tanh_tab} show the numerical results.
\subsubsection{Trying different window sizes/sequence length}
\begin{figure}[H]
    \centering
    \includegraphics[width = 0.7\linewidth]{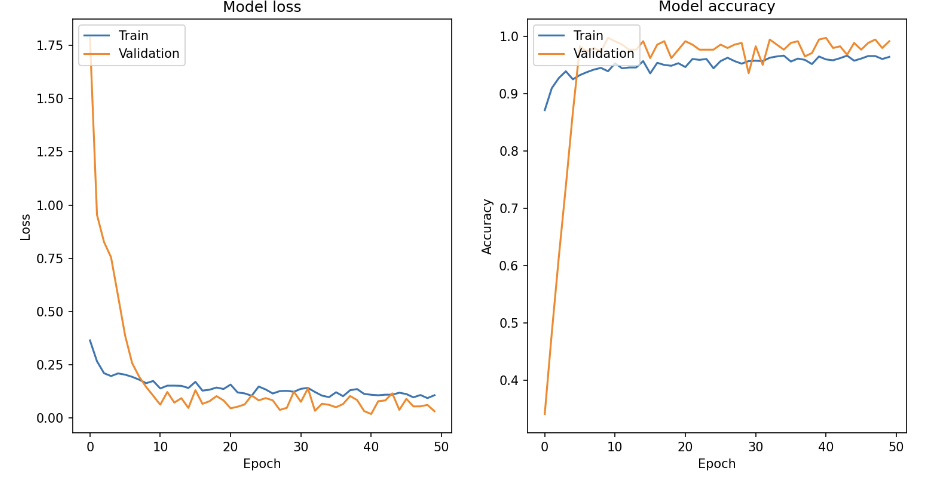}
    \caption{accuracy and loss  of training and validation datasets across epochs of using a window size =25}
    \label{window_size}
\end{figure}

 In this paper, we tried 2 window sizes which are 20 (base model) and 25 .By comparing results of the test datasets, it is shown that the window size of 20 showed better results than window size of 25. Fig. \ref{window_size} shows the graph results and Table \ref{window_size_tab} shows the numerical results.
\vspace{-10pt}\vspace{-10pt}
\vspace{-10pt}\begin{table}[H]
    \centering
    \caption{Results of using sigmoid as an activation function  on test dataset}
    \label{sigmoid_tab}
    \begin{tabular}{|c|c|c|c|c|}
        \hline
         Class& Precision & Recall &  F1-score & support \\
        \hline
        Going Straight &   0.86 &      0.95  &    0.90 &       719 \\
        \hline
        Turn Left & 0.93   &   0.92    &  0.92 &      857 \\
        \hline
        Turn Right &    0.96 &     0.82  &    0.88   &    486 \\
        \hline
    \end{tabular}
\end{table}\vspace{-7pt}
\vspace{-10pt}\vspace{-10pt}
\vspace{-10pt}\begin{table}[H]
    \centering
    \caption{Results of using tanh as an activation function  on test dataset}
    \label{tanh_tab}
    \begin{tabular}{|c|c|c|c|c|}
        \hline
         Class& Precision & Recall &  F1-score & support \\
        \hline
        Going Straight &   0.86 &     0.96  &    0.91   &    719\\
        \hline
        Turn Left & 0.96   &   0.91   &   0.93      & 857 \\
        \hline
        Turn Right &    0.93 &     0.84  &    0.89    &   486\\
        \hline
    \end{tabular}
\end{table}\vspace{-7pt}
\vspace{-10pt}\vspace{-10pt}
\vspace{-10pt}\begin{table}[H]
    \centering
    \caption{Results of using a window size =25}
    \label{window_size_tab}
    \begin{tabular}{|c|c|c|c|c|}
        \hline
         Class& Precision & Recall &  F1-score & support \\
        \hline
        Going Straight &  0.81 &     0.97  &    0.88   &    719\\
        \hline
        Turn Left & 0.96   &   0.85   &   0.90  &    857 \\
        \hline
        Turn Right &    0.94   &   0.85   &   0.89 &       486 \\
        \hline
    \end{tabular}
\end{table}\vspace{-7pt}
 \subsubsection{Ablation studies: trying different number of hidden layers }
In relevance to our basic model which had 3 hidden layers where they all have relu activation functions , we investigate 2 hidden layers and 4 hidden layers thus we once try removing a layer and once adding a layer. The results stated   that our base model is the best , then the  2  hidden layers and lastly the 4 hidden layers.
\vspace{-10pt}\begin{table}[H]
    \centering
    \caption{Results of using only 2 hidden layers}
    \label{2 hidden}
    \begin{tabular}{|c|c|c|c|c|}
        \hline
         Class& Precision & Recall &  F1-score & support \\
        \hline
        Going Straight &  0.87    &  0.97   &   0.92  &     719\\
        \hline
        Turn Left & 0.96   & 0.84   &   0.89 &      857\\
        \hline
        Turn Right & 0.83   &   0.86   &   0.84 &      486\\
        \hline
    \end{tabular}
\end{table}\vspace{-7pt}
\vspace{-10pt}\begin{table}[H]
    \centering
    \caption{Results of using  4 hidden layers}
    \label{4 hidden}
    \begin{tabular}{|c|c|c|c|c|}
        \hline
         Class& Precision & Recall &  F1-score & support \\
        \hline
        Going Straight &   0.84  &    0.92  &    0.88    &   719\\
        \hline
        Turn Left & 0.90   &   0.81  &    0.86  &     857\\
        \hline
        Turn Right & 0.80 &     0.84  &    0.82 &      486\\
        \hline
    \end{tabular}
\end{table}\vspace{-7pt}
\subsubsection{Tackling different learning rates}
A learning rate of  0.01 is tested and by comparison of test data set results , the 0.001 learning rate (base model )yielded way better results than the 0.01 .
\vspace{-10pt}\begin{table}[H]
    \centering
    \caption{Results of using  an 0.01 learning rate }
    \label{lr}
    \begin{tabular}{|c|c|c|c|c|}
        \hline
         Class& Precision & Recall &  F1-score & support \\
        \hline
        Going Straight &   0.96   &   0.22  &    0.36     &  719\\
        \hline
        Turn Left &  0.62   &   0.62  &    0.62 &      857\\
        \hline
        Turn Right & 0.45   &   0.96  &    0.61 &      486\\
        \hline
    \end{tabular}
\end{table}\vspace{-7pt}
\subsubsection{Prediction window}
In our base model , the predictions were 2 seconds earlier than the real event (base model) but we also opted to try the %how would the model be affected if we predicted 
prediction only 1.5 seconds earlier.The outcome is that predicting 2  s earlier yielded  better results.
\vspace{-10pt}\begin{table}[H]
    \centering
    \caption{Results of using  a 1.5 sec  prediction window  }
    \label{pw}
    \begin{tabular}{|c|c|c|c|c|}
        \hline
         Class& Precision & Recall &  F1-score & support \\
        \hline
        Going Straight &  0.85    &  0.97   &   0.91  &     719\\
        \hline
        Turn Left &  0.95    &  0.82  &    0.88 &       857\\
        \hline
        Turn Right & 0.80   &   0.83  &    0.81 &       486
\\
        \hline
    \end{tabular}
\end{table}\vspace{-7pt}
It was witnessed  in the aforementioned variations that tuning a single parameter  exceedingly affects the model, so to reach the best model plenty of trial and error has to be carried and in our experiments the base model is the best one, but this should not hold true for different data sets.Further more, it was alternating  a single factor and measuring the performance but also other  maximal performances could be obtained if more than one factor are changed at a time.

\section{Conclusion and Future Recommendations}\label{sec7}

In this paper, LSTM has been proven exceedingly effective and promising in solving the classification problem of intention prediction and the most outstanding model in our work is the base model yielding a maximum training accuracy of 97.24\%  and a maximum validation accuracy of 99.71\%. 
As future recommendations, it is aimed that our model is utilized in intersection management and further tested on other environments and driving scenarios.

%In extension of our efforts in this paper , it is aimed that we test our model on  synthetic data from CARLA simulator by extracting the same exact features using sensors and testing the applicability on CARLA which offers an ego vehicle view , the model is expected to   work perfectly fine because it is only dependent on numerical features thus verification of this hypothesis should be verified.\par%

%Additionally , our paper could serve as a promising starting point to be used in vehicle to infrastructure (v2x) because BEV was used so if view is captured from infrastructure then the infrastructure could communicate with vehicles to convey the intentions of leading vehicles because the infrastructure then has the full overview and vehicle interactions.%
%Not only is it aimed to  continue improving our LSTM model but we are also looking forward to trying other new deep learning  architectures like transformers and compare them to LSTM  in tackling the problem of vehicle intention  prediction.Finally, we would like to generalise and explore other driving scenarios so it would be highly promising if we test it on roundabouts and even alter our model accordingly.%

\appendices


%\input{sections/ack}
\bibliographystyle{IEEEtran}
\bibliography{sections/ref} %edit all the generated bibtex from google scholar in sections/ref.bib
\nocite{manzour2023vehicle}
\nocite{laimona2020implementation}
\nocite{zhang2021vehicle}
\nocite{zhu2023multi}
\nocite{sanchez2022foresee}
\nocite{li2021open}
\nocite{girma2020deep}
\nocite{cao2021real}
\nocite{yan2023int2}
\nocite{el2021effective}
\nocite{qiao2020human}
\nocite{mozaffari2020deep}
\nocite{phillips2017generalizable}

\end{document}